\documentclass[10pt,twocolumn,letterpaper]{article}

\usepackage[preprint]{cvpr}      
\usepackage{algorithm}
\usepackage{algorithmic}
\usepackage{multicol}
\usepackage{booktabs}
\usepackage{pifont}
\usepackage{multirow}
\usepackage{amsmath}

\definecolor{cvprblue}{rgb}{0.21,0.49,0.74}
\usepackage[pagebackref,breaklinks,colorlinks,allcolors=cvprblue]{hyperref}

\def\paperID{6448} 
\def\confName{CVPR}
\def\confYear{2025}

\title{IE-Bench: Advancing the Measurement of Text-Driven Image Editing for Human Perception Alignment}

\author{Shangkun Sun\textsuperscript{\rm 1, \rm 2}, Bowen Qu\textsuperscript{\rm 1}, Xiaoyu Liang\textsuperscript{\rm 1}, Songlin Fan\textsuperscript{\rm 1}, Wei Gao\textsuperscript{\rm 1, \rm 2}\thanks{Corrsponding author} \\
Peking University\textsuperscript{\rm 1}, PengCheng Laboratary\textsuperscript{\rm 2}
}

\begin{document}
\maketitle
\begin{abstract}
Recent advances in text-driven image editing have been significant, yet the task of accurately evaluating these edited images continues to pose a considerable challenge.
Different from the assessment of text-driven image generation, text-driven image editing is characterized by simultaneously conditioning on both text and a source image. The edited images often retain an intrinsic connection to the original image, which dynamically change with the semantics of the text. However, previous methods tend to solely focus on text-image alignment or have not aligned with human perception. In this work, we introduce the Text-driven Image Editing Benchmark suite (IE-Bench) to enhance the assessment of text-driven edited images. IE-Bench includes a database contains diverse source images, various editing prompts and the corresponding results different editing methods, and total 3,010 Mean Opinion Scores (MOS) provided by 25 human subjects. Furthermore, we introduce IE-QA, a multi-modality source-aware quality assessment method for text-driven image editing. To the best of our knowledge, IE-Bench offers the first IQA dataset and model tailored for text-driven image editing. Extensive experiments demonstrate IE-QA's superior subjective-alignments on the text-driven image editing task compared with previous metrics. We will make all related data and code available to the public.
\end{abstract}    
\section{Introduction}
\label{sec:intro}

\begin{figure*}[t]
\centering
\includegraphics[width=1.5\columnwidth]{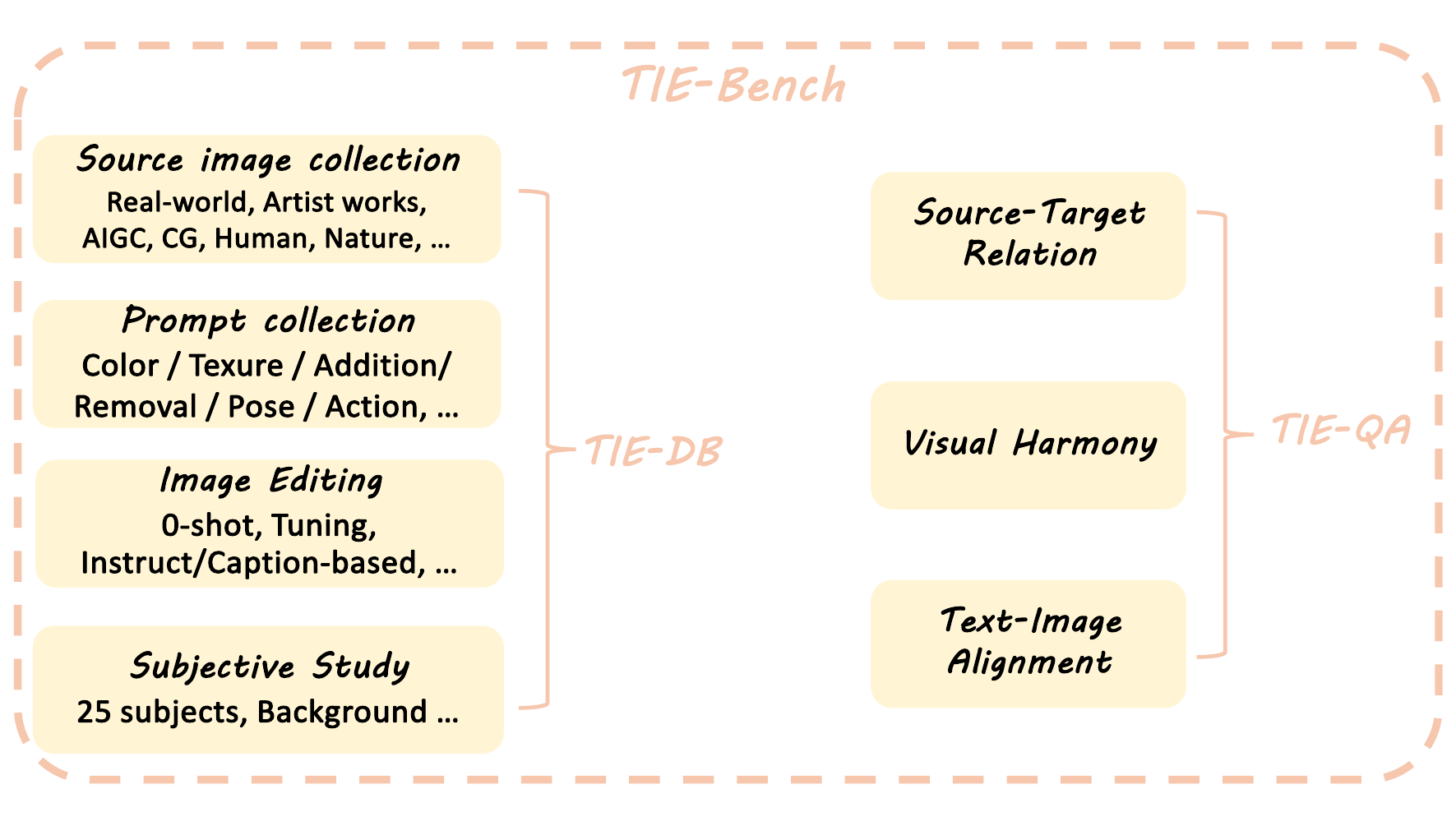} 
\caption{Overview of the proposed IE-Bench.}
\label{fig:1}
\end{figure*}

Text-driven image editing~\cite{instructpix2pix, masactrl, magicbrush, sine, dac} has attracted significant attention in recent years. However, there is currently no well-established metric for evaluating the results of image editing. Objective metrics such as CLIP score~\cite{clip}, DINO score~\cite{dino}, LPIPS score~\cite{lpips}, and SSIM~\cite{ssim} tend to evaluate image quality from a single perspective, such as text-image consistency or the correlation between the pre- and post-editing images. These metrics, however, do not provide an overall evaluation, nor do they align well with human visual perception. Previous studies~\cite{t2iqa, hps, pickscore} have shown that these metrics can significantly differ from human judgment in practical applications.

In recent years, some metrics aligned with human perception, such as HPS scores~\cite{hps, hpsv2}, Pick score~\cite{pickscore}, and ImageReward~\cite{imagereward}, have made effective progress in evaluating text-to-image generation tasks by collecting human visual feedback. However, these methods focus only on individual images and text, which differs from the setting of image editing tasks. Unlike text-driven image generation, text-driven image editing also takes a source image as input. The edited result is expected to differ from the source image, but there is also a certain degree of correspondence. Modeling this relationship is crucial for evaluating the editing result: in some cases, the edited result is expected to retain semantic information related to the original image. If only the edited image output is considered, this aspect would be missed. This is an issue that previous methods have not addressed in depth.

However, modeling this relationship is a challenging task. The connection between source and target images changes dynamically depending on the text context. For example, a stylistic instruction like "make it a claymation style" may drastically alter the structure, texture, and lines of the original image, whereas a replacement instruction like "replace the cat with a dog" will directly alter the semantic content, and thus a large difference between the source and target images is expected. On the other hand, an instruction like "remove her earrings" is expected to retain most of the identity information of the original character. Therefore, a multi-modal method that can dynamically model the source-target relationship is urgently needed.

In this work, we propose the Text-driven Image Editing Benchmark (IE-Bench) to improve the alignment between evaluation metrics for text-driven image editing and human perception. We first introduce IE-DB, a database containing various source-prompt-target cases and their corresponding Mean Opinion Scores (MOS). We collect diverse real-world, CG, AIGC, and art painting images from different sources, as shown in Figure~\ref{fig:src_image_collection}. Following previous works~\cite{huang}, we manually design diverse editing instructions for each image, covering aspects such as structural changes (e.g., shape, size), style changes (e.g., texture, color), and semantic changes (e.g., pose, action, addition, replacement, deletion). We then apply multiple methods to generate diverse edited results. Finally, we assemble 25 human participants from various backgrounds to provide subjective ratings. These ratings follow the ITU standard~\cite{itu}, and the detailed process is described in Section~\ref{sec:tie_data}. To the best of our knowledge, this is the first image quality assessment dataset specifically for image editing.

Building on IE-DB, we further develop IE-QA, a source-aware multi-modal assessment method for text-driven image editing. IE-QA introduces an effective way to model the source-target relationship in image editing tasks and provides a comprehensive evaluation by considering multiple dimensions, such as the quality of the edited image, the relationship between the image and text, and the connection between the source and target images. Extensive experiments demonstrate the effectiveness of IE-QA and its superior alignment with human perception compared to traditional methods.

Our contributions are summarized as follows:
\begin{itemize}
    \item We introduce IE-DB, which collects diverse source-prompt-target cases from various sources along with corresponding MOS reflecting human subjective evaluations. To the best of our knowledge, this is the first image quality assessment dataset for text-driven image editing.
    \item We further propose IE-QA, which incorporates both the source-target relationship and multi-modal text input, offering an effective way to dynamically model the source-target relationship in response to multi-modal inputs.
    \item IE-QA demonstrates a significant advantage in subjective alignment compared to previous IQA methods, effectively showcasing the potential of modeling the source-target relationship during evaluation.
\end{itemize}

\section{Related Work}
\label{sec:related_work}

\subsection{Image Quality Assessment}

In recent years, researchers have introduced numerous image quality assessment (IQA) methods~\cite{agiqa,clipiqa,dbcnn,liqe,linearityiqa,cnniqa,hyperiqa,IP-IQA}. These methods can be categorized into full-reference IQA (FR-IQA)~\cite{tang2018full, zhang2012comprehensive, larson2010most} and no-reference IQA (NR-IQA)~\cite{clipiqa,dbcnn,liqe,linearityiqa}, based on the presence or absence of a reference image during the assessment process. Full-reference methods generally demonstrate higher prediction accuracy than no-reference methods, as a reference image enables the extraction of more effective features. Many classical IQA models initially utilized manually extracted feature-based methods~\cite{larson2010most, zhang2012comprehensive}. However, with the rapid advancement of convolutional neural networks, methods that employ deep learning for feature extraction~\cite{liqe,linearityiqa,cnniqa,hyperiqa,IP-IQA} have significantly enhanced performance. ~\cite{kang2014convolutional} is the early adopter of deep convolutional neural networks for no-reference image quality assessment (NRIQA), employing a CNN to derive image quality metrics directly from raw image data instead of relying on traditional hand-crafted features or a reference image. ~\cite{dbcnn} pioneered the use of a deep bilinear convolutional neural network specifically for blind image quality assessment, merging two CNN pathways to separately evaluate synthetic and natural image distortions. Meanwhile, ~\cite{hyperiqa} developed a self-adaptive hyper network, which innovatively determines the quality of images with authentic distortions using a methodical three-stage approach: analyzing the content, learning perception rules, and predicting image quality. However, these methods often do not account for the differing source-target relationships caused by changes in multi-modal contexts.

\subsection{Metrics for Image Editing}

Currently, the evaluation methods commonly used in text-driven image editing include several objective metrics~\cite{clip,lpips,ssim,fid}, as well as some IQA methods~\cite{hps,hpsv2,imagereward,pickscore} aligned with human feedback. CLIP-V~\cite{clip} calculates the cosine similarity between each edited image and the source image, while CLIP-T measures the relationship between the result and the given text prompt. MSE and SSIM~\cite{ssim} represent the variance in pixels and overall structure between the edited image and the source image. FID~\cite{fid} calculates the Fréchet Distance between two images. The DINO score~\cite{dino} measures semantic consistency and calculates feature variance. However, these individual metrics often assess the editing results only from a single dimension. For instance, they either measure only the source-target relationship or the visual-text connection, without aligning with human subjective perception. PickScore~\cite{pickscore} estimates alignment with human preferences via a CLIP-style model fine-tuned on human preference data. ImageReward~\cite{imagereward}, and HPS scores~\cite{hps, hpsv2}, evaluate natural images from aesthetic and technical distortion perspectives. Despite effective scoring based on human feedback training, these methods do not consider the inherent relationship between edited results and the source image, and some traditional IQA methods like~\cite{dbcnn} do not model the alignment between text and image. Currently, there is still a lack of an appropriate metric to evaluate edited results based on the source image and the editing prompts.

\subsection{Datasets for Image Editing}

In previous studies, a common approach to evaluating edited images was to assemble human annotators to conduct subjective preference experiments~\cite{smartedit, refcoco-edit}. However, the results of such subjective experiments are often difficult to reproduce, and there can be significant variability in the data and prompts selected when comparing different methods. Recently, some research~\cite{instructpix2pix, sine} has attempted to address this by curating high-quality image editing prompt pairs through diverse data collection and prompt design, aiming to standardize community assessments. Nevertheless, these efforts still face two challenges:

1. These datasets do not include subjective experimental feedback corresponding to the image data (such as mean opinion scores, MOS). Therefore, when using these datasets, others may still need to rely on objective metrics or conduct new subjective experiments.

2. The scenarios covered by these datasets could potentially be expanded further.

\subsection{Methods for Image Editing}
Pre-trained text-to-image diffusion models~\cite{dreambooth, sd, ddpm} have been proven to be highly effective in image editing tasks. Instruction-based image editing~\cite{instructpix2pix, magicbrush} requires users to provide an instruction to transform the original image into a new one that conforms to the given instruction. Some methods can achieve this goal without fine-tuning. For example, Prompt-to-Prompt~\cite{pnp} suggests modifying cross-attention maps by comparing the descriptions of the original input with those of the modified version. MasaCtrl~\cite{masactrl} transforms the self-attention mechanism within diffusion models into mutual self-attention, allowing the model to query relevant local content and textures from the source image, thereby enhancing consistency. Furthermore, due to the scarcity of paired image-instruction editing datasets, pioneering work InstructPix2Pix~\cite{instructpix2pix} introduced a large-scale vision-language dataset created by fine-tuning GPT-3~\cite{gpt3} and Prompt-to-Prompt~\cite{pnp} with stable diffusion. This method further fine-tuned the UNet architecture to enable the model to edit images based on simple instructions. To improve the editing performance of InstructPix2Pix on real-world images, MagicBrush~\cite{magicbrush} provided a large-scale manually annotated dataset for instruction-guided real image editing. SmartEdit~\cite{smartedit} leverages large language models (LLMs) to enhance text understanding during the editing process, while SINE~\cite{sine} and DAC~\cite{dac} further boost the model's representation ability for specific samples through test-time fine-tuning. However, these methods typically rely on objective metrics~\cite{clip, dino} for evaluation, which do not fully align with human subjective perception or only conduct one-off subjective experiments, leading to poor comparability and reproducibility. There remains a lack of comprehensive evaluation metrics that align well with human subjective feedback.

\section{Text-driven Image Editing Database}
\label{sec:tie_data}
The collection of IE-DB involves four primary stages: source image collection, prompt selection and execution of image editing methods, and subjective experiments conduction. We will elaborate on each part in subsequent sections with corresponding analyses.

\begin{figure*}[t]
\centering
\includegraphics[width=2.0\columnwidth]{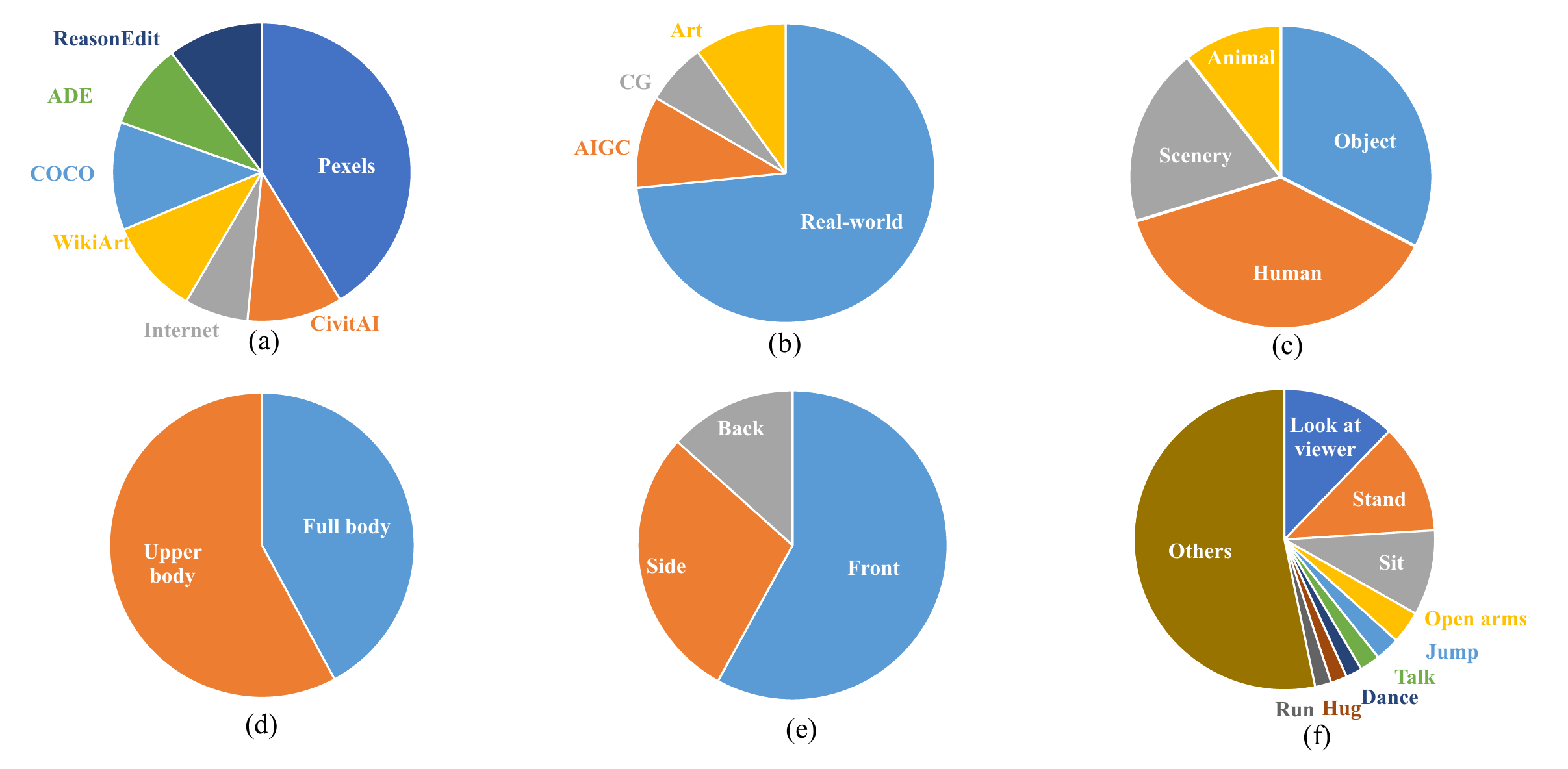} 
\caption{Collection of source images. (a) Sources of images. (b) Categories of images. (c) Content of images. (d)~(f) denotes fine-grained classification of human images, including classification of camera, pose, and actions, respectively.} 
\label{fig:src_image_collection}
\end{figure*}

\subsection{Source Image Collection}
IE-DB compiles a rich and varied collection of source images to facilitate a more robust evaluation of image editing quality. In addition to real-world scenes, the dataset incorporates computer-generated imagery, text-driven generated images, and artistic creations. Real-world scenes, given their widespread application, constitute the largest portion of the dataset. Unlike previous efforts, IE-DB refrains from random sampling of images from large datasets to avoid complications related to copyright, watermarks, and image resolution. Instead, to cover as many different content subjects, action categories, and scenarios as possible, IE-DB manually selected 301 images from four datasets: ADE~\cite{ade}, WIKIArt~\cite{wikiart}, COCO~\cite{coco}, ReasonEdit~\cite{smartedit}, and other Internet sources. Instead of sampling from large-scale dataset, We first confirmed the data sources (CG, AIGC, real-world datasets, and artist-created works), then iterated through each sample to tag attributes such as ``[Landscape/Object/Animal/Human]" and ``[action type]". Samples of the same type (e.g., Landscape/Object/Animal or Human action) or with insufficient resolution were skipped until the dataset was fully reviewed or the relevant categories were sufficiently populated. Smaller datasets were prioritized for review, and additional samples were drawn from larger datasets. Finally, suitable images were selected from the internet for supplementation. For instance, while many datasets included landscapes like grasslands and snowy mountains, scenes such as auroras, lava flows, and lightning were less represented. Similarly, although current datasets provide a rich variety of action categories, there remains a limited number of actions with significantly varied patterns. Ultimately, we collected over 100 human action types, as illustrated in the Figure~\ref{fig:src_image_collection}. For simplicity, the remaining categories were grouped under "others," with detailed explanations provided in the supplementary materials. Ultimately, a diverse set of 301 source images was gathered, encompassing a wide range of content. The origins, contents, and category distributions of these images are illustrated in Figure~\ref{fig:src_image_collection}. Each image was resized to ensure the shorter side was 512 pixels, preserving the original aspect ratios.

\subsection{Prompt Selection}
\begin{figure*}[htbp]
\centering
\includegraphics[width=1.8\columnwidth]{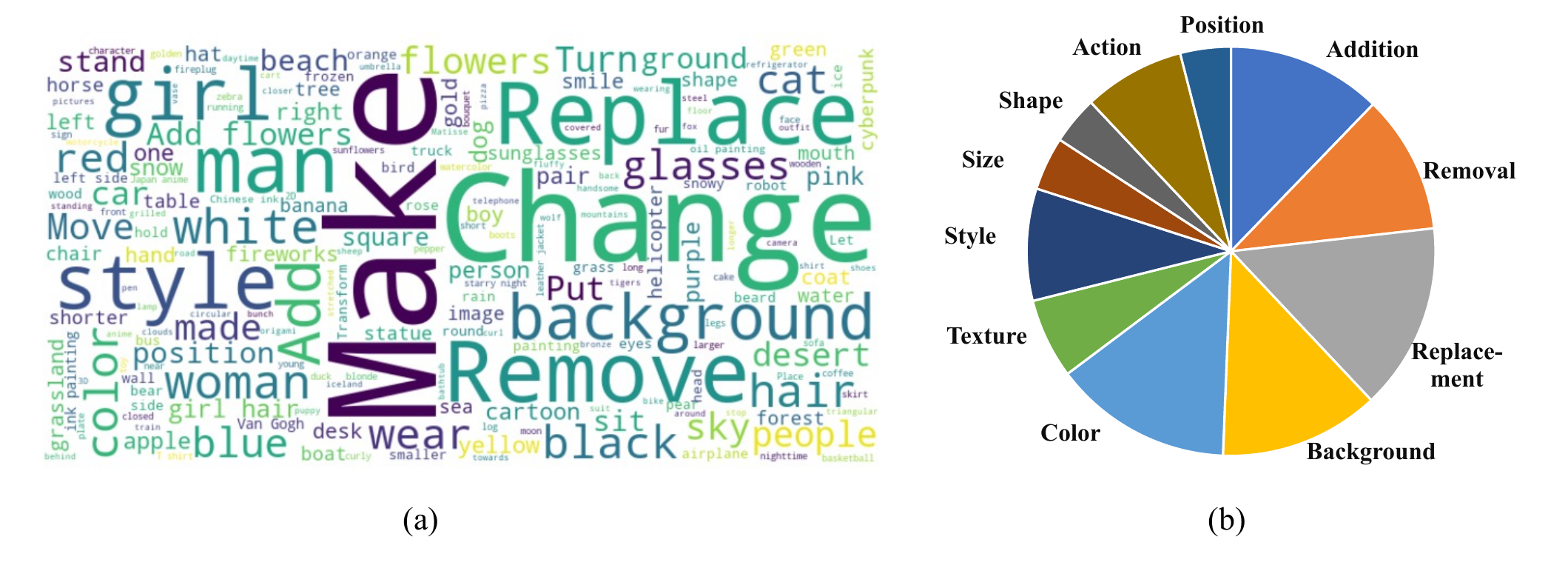} 
\caption{Statistics of IE-DB prompts. (a) Word cloud of IE-Bench DB prompts. (b) Proportion of different types}

\label{fig:prompt_info}
\end{figure*}

Building upon previous research~\cite{prompt}, we categorize image editing prompts into three primary types: (1) Style editing, involving adjustments to color, texture, or overall ambiance. (2) Semantic editing, which encompasses background modifications and localized edits such as adding, replacing, or removing specific objects. (3) Structural editing, including changes to object size, pose, or motion. To guarantee the variety and specificity of the prompts, we created tailored prompts for each image, with the distribution detailed in Figure~\ref{fig:prompt_info}.

\subsection{Image Editing}
We selected five diverse image editing techniques. To achieve a balanced quality distribution among the edited images, our selection includes both state-of-the-art models and earlier methods. Additionally, we incorporated approaches based on various foundational models, from SD 1-4 to SD2-1, to enhance the diversity of the editing outcomes.
Moreover, to diversify the edited content, we included both zero-shot methods and those that necessitate fine-tuning. We also opted for models that employ different editing paradigms, such as Instruct-P2P~\cite{instructpix2pix}, Prompt-to-prompt~\cite{pnp}, and MasaCtrl~\cite{masactrl}, among others. The specifics of these methods are outlined in Table~\ref{tab:editing_methods}.

\begin{table*}[htb]
\centering
\scalebox{1.0}{
\begin{tabular}{llccc}
\toprule
Model & Time & 0-shot & Type & SD. \\ \midrule
Instruct-Pix2Pix~\cite{instructpix2pix}      &  CVPR'23   &    \ding{55}   &   Instruction-based  &    1-4       \\ \midrule
Prompt-to-Prompt~\cite{pnp}      &  CVPR'23    &    \ding{51}       &     Description-based       &   1-5         \\ \midrule
MagicBrush~\cite{magicbrush}      &  NeurIPS'23    &    \ding{55}       &     Instruction-based       &   1-5    \\ \midrule
MasaCtrl~\cite{masactrl}      & ICCV'23     &   \ding{51}          &    Description-based             &  1-4          \\ \midrule
InfEdit~\cite{infedit}      & CVPR'24     &  \ding{51}           &  Description-based      &    2-1       \\
\bottomrule
\end{tabular}
}
\caption{Collection of the editing models.}
\label{tab:editing_methods}
\end{table*}

\subsection{Subjective Study}
In accordance with ITU standards~\cite{itu}, subjective experiments necessitate a minimum of 15 participants to ensure result variance remains within acceptable limits. For this study, we enlisted 25 participants with diverse professional backgrounds. These individuals were tasked with evaluating edited images based on text-image consistency, source-target fidelity, and overall quality, relying on their subjective judgments.

All participants were over 18 years old, held at least an undergraduate degree, and had varied professional experiences, including business, engineering, science, and law, ensuring their ability to make independent judgments. Prior to the experiment, participants underwent in-person training, where they were shown examples of high-quality and poor edits not included in the dataset. During the assessment, each participant evaluated all image samples, with a mandatory 5-minute break every 15 minutes to minimize fatigue. Our procedures were consistent with those in previous subjective studies, such as~\cite{t2vqa, agiqa, keimel2012tum}.

Text-image consistency was defined as the extent to which the edited content aligns with the provided prompt. Source-target fidelity measures how well the edited image retains a connection to the original. Participants rated these aspects on a 1 to 10 scale during their evaluations.

\begin{figure}[t]
\centering
\includegraphics[width=0.8\columnwidth]{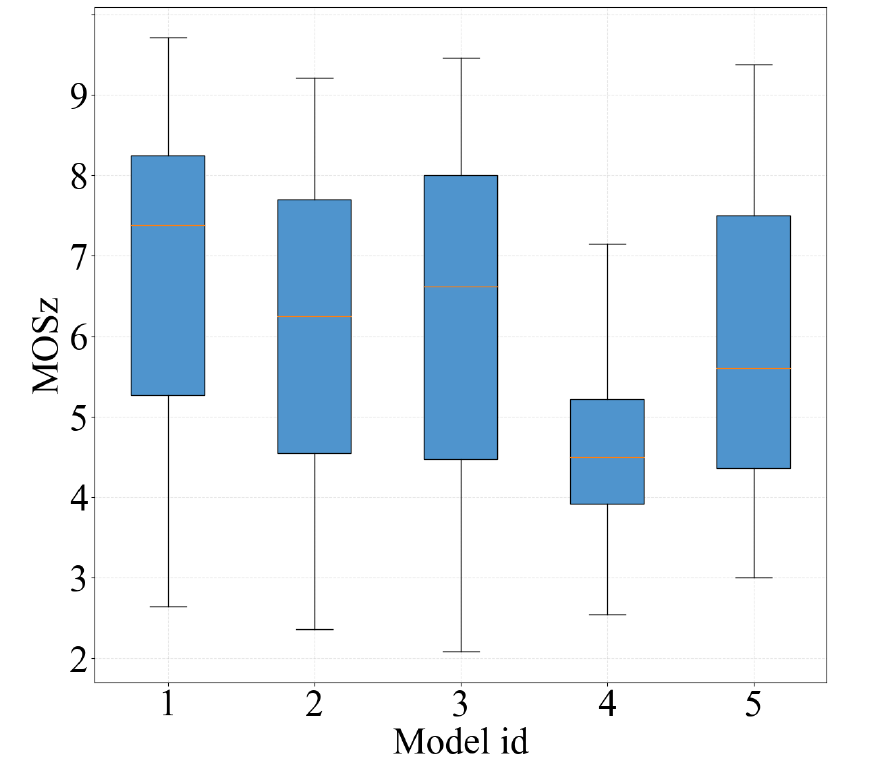} 
\caption{Z-score MOS distributions of different editing methods.}

\label{fig:mos}
\end{figure}

Consistent with prior research~\cite{e-bench,t2vqa,liqe}, we employed Z-score normalization for the raw MOS values. Following the collection of raw Mean Opinion Scores (MOS), we applied Z-score normalization to account for inter-subject variability, using the formula:

\begin{align}
    Z_{m,i} = \frac{X_{m,i} - \mu(X_i)}{\sigma(X_i)},
\end{align}

where $X_{m,i}$ denotes the raw MOS and $Z_{m,i}$ the Z-score for the $m$-th image evaluated by the $i$-th participant. Here, $\mu(\cdot)$ and $\sigma{(\cdot)}$ represent the mean and standard deviation, respectively, and $X_{i}$ is the set of all MOS scores from participant $i$. We also utilized the outlier filtering method from BT.500~\cite{bt500}. Figure~\ref{fig:mos} illustrates the distinction between raw and normalized scores.

\subsection{Statistcs Analysis}
We performed a comprehensive analysis of the IE-DB dataset and examined the evaluation scores for the image editing outcomes produced by each model on the gathered images, as depicted in Fig.~\ref{fig:mos}. The models, labeled from 1 to 5, are MagicBrush~\cite{magicbrush}, Instruct-Pix2Pix~\cite{instructpix2pix}, Inf-Edit~\cite{infedit}, MasaCtrl~\cite{masactrl}, and Prompt-to-Prompt~\cite{pnp}. Our observations reveal that the median scores for these models range between 4 and 8, with most scores clustering around 6. Notably, some edits received scores as low as 2, while others achieved scores above 9. This variability suggests that there is significant potential for enhancing the models' performance. The observed limitations may stem from the models' inadequate comprehension of prompts and their limited capability to capture fine visual details.
\section{Method}

\subsection{Text-driven Image Editing Quality Assessment}

\begin{figure}[t]
\centering
\includegraphics[width=1.05\columnwidth]{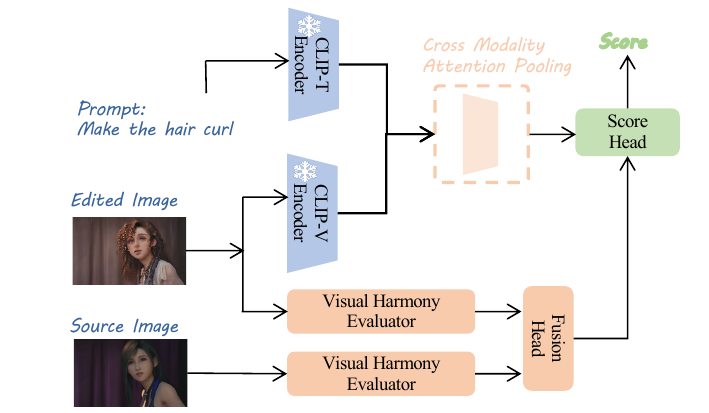} 
\caption{Network architecture of IE-QA.}
\label{fig:network}
\end{figure}

Utilizing the IE-DB, we developed the IE Quality Assessment (IE-QA) network, designed to mirror human subjective evaluations in assessing the quality of edited images, as depicted in Figure~\ref{fig:network}. This network evaluates edited images based on three key aspects: image-text alignment, source-target relationship, and visual quality.

\subsection{image-Text Alignment}

Conventional image evaluation techniques often overlook the alignment between image content and the associated text prompt, a limitation that hampers their effectiveness in assessing AI-generated imagery. To address this, we integrated a text branch into our model, inspired by a successful IQA approach, to capture the alignment between the generated content and the corresponding text prompt. This can be expressed as:

\begin{align}
    e_{bv} &= \mathbf{F_{bv}}(I^\star), \\
    e_{bt} &= \mathbf{F_{bt}}(p, \mathbf{F_{ca}}(e_{bv}))
\end{align}

Here, $\mathbf{F_{bv}}$ and $\mathbf{F_{bt}}$ represent the CLIP visual and text encoders, respectively. $p$ denotes the prompt, and the visual feature $e_{bv}$ is interacted with the text encoder through cross-attention, denoted as $\mathbf{F_{ca}}$.

\subsection{Source-Target Relationship}
Assessing the consistency between original and edited images is inherently complex due to their latent connections and pronounced pixel-level discrepancies. To tackle this, we developed a multi-modal feature extractor that projects both the original and edited images into a latent space. By concatenating these features and processing them through a feed-forward network, we obtain a reliable measure of their relevance. This process is formulated as:

\begin{align}
    f &= \mathbf{F}(I), \\
    f^\star &= \mathbf{F^\star}(I^\star), \\
    o_s &= \mathbf{H_s}(Concat(f, f^\star))
\end{align}

Here, $\mathbf{F}$ and $\mathbf{F^\star}$ represent the pre-trained visual encoders for the original and edited images, respectively. The output vector $o_s$ captures the relevance between the source and edited images, with $\mathbf{H_s}$ denoting the lightweight feed-forward network. In practice, we experimented with various backbones and ultimately selected the CLIP visual encoder for its effectiveness.

\subsection{Visual Quality}
To evaluate the quality of the edited image, we adopt perspectives from the top-performing IP-IQA method~\cite{IP-IQA}, which assesses images based on aesthetics and technical distortion. Initially, the pre-trained visual encoder's parameters are frozen, and only the regression head is fine-tuned. In the subsequent stage, all parameters of the visual quality branch are updated to refine the assessment.

\subsection{Supervision}
Following previous research, we employ a combined loss function comprising the Pearson Linear Correlation Coefficient (PLCC) loss and rank loss, weighted by $\alpha$, to train the entire network. This comprehensive loss function is formulated as:
\begin{align}
    L &= L_{plcc} + \alpha \cdot L_{rank},
\end{align}
where $\alpha$ is set to 0.3 in practice.

\section{Experiments}

\subsection{Implementation Details}
All models were developed using PyTorch and trained on NVIDIA V100 GPUs. Following the 10-fold cross-validation approach used in previous studies~\cite{hyperiqa,IP-IQA,dbcnn}, all models were trained on the IE DB dataset with an initial learning rate of 1e-3, a batch size of 8, and a total of 60 epochs. Building upon the approach outlined in IP-IQA~\cite{IP-IQA}, we initially fine-tuned the model head using linear probing for 40 epochs, followed by training all model parameters for an additional 20 epochs. During training, the Adam~\cite{adam} optimizer was employed in conjunction with a cosine learning rate scheduler. In line with previous research, the base visual-text interaction block utilizes attention pooling, leveraging the CLIP visual and text encoders.

\subsection{Evaluation Metrics}
Consistent with prior studies~\cite{IP-IQA,t2vqa,dbcnn}, we use four metrics as our evaluation metrics: Spearman’s Rank Order Correlation Coefficient (SROCC), Pearson’s Linear Correlation Coefficient (PLCC), Kendall rank-order correlation coefficient (KRCC). 

\subsection{Quantitative Results}
Our results were compared against advanced evaluation metrics in image editing, including objective methods~\cite{clip,fid, hps, hpsv2,pickscore} and human-aligned image Quality Assessment (VQA) methods~\cite{dbcnn, hyperiqa, IP-IQA}, as shown in Table~\ref{tab:tie_bench}.
From these, it can be seen that compared to previous traditional VQA methods~\cite{IP-IQA, hyperiqa, dbcnn} and commonly used objective metrics, IE-QA outperforms previous traditional VQA methods and commonly used objective metrics in aligning with human subjective perception, achieving improvements of $10.46\%$, $9.02\%$, $10.19\%$, $11.08\%$ in SROCC, PLCC, KLCC, and RMSE, respectively. 
Compared to the baseline, IE shows significant performance gains, with increases 0.1396, 0.1435, 0.096, and 0.243 in SROCC, PLCC, KLCC, and RMSE, respectively. Among the 0-shot objective measurements, performance tends to be lower compared to learning-based methods, a trend observed in previous studies such as~\cite{fid,dino}. This is likely due to the difficulty in aligning objective quantitative metrics with human perceptions. 

Among the 0-shot methods, HPSv2~\cite{hpsv2}, which is pre-trained with a reward model incorporating human feedback, achieved the highest overall performance. CLIP-V~\cite{clip} showed comparable SROCC and KRCC metrics to CLIP-T but failed to align well with human subjective perception, resulting in lower scores.

\begin{table*}[htb]
\centering
\scalebox{1.0}{
\begin{tabular}{llcccc}
\toprule
\multirow{2}{*}{Type} & \multirow{2}{*}{Models} & \multicolumn{4}{c}{IE DB 10-fold} \\ \cmidrule{3-6}
& & SROCC $\uparrow$ & PLCC $\uparrow$ & KRCC $\uparrow$ & RMSE $\downarrow$ \\ \midrule
\multirow{6}{*}{Subjective Metrics}
&CLIP-T~\cite{clip} & 0.1503 & 0.1229 & 0.1487 & 4.645 \\
&CLIP-V~\cite{clip} & 0.1672 & 0.1405 & 0.0954 & 4.766 \\
&DINO Score~\cite{dino} & 0.1187 & 0.0639 & 0.0817 & 5.4554 \\ 
&LPIPS~\cite{lpips} & 0.0159 & 0.0425 & 0.0327 & 5.7034 \\
&SSIM~\cite{ssim} & 0.0165 & 0.0194 & 0.0173 & 4.7180 \\
&PSNR & 0.0923 & 0.1547 & 0.1073 & 6.1025 \\
\midrule
\multirow{7}{*}{Human-Aligned Metrics}
&HPS-v2~\cite{hpsv2} & 0.2111 & 0.2153 & 0.1471 & 4.751 \\
&Pick-a-Pic~\cite{pickscore} & 0.1734 & 0.1673  & 0.1138  & 4.8063 \\
&ImageReward~\cite{imagereward} & 0.1852 & 0.2055 & 0.1402 & 2.2483 \\
& DBCNN~\cite{dbcnn} & 0.4161 & 0.3507 & 0.2832 & 4.113 \\
& HyperIQA~\cite{hyperiqa} & 0.3302 & 0.3916 & 0.366 & 4.793 \\
& IP-IQA~\cite{IP-IQA} & 0.6124 & 0.6063 & 0.4581 & 1.288 \\
\midrule
Ours
& IE-QA & \textbf{0.7520} & \textbf{0.7498} & \textbf{0.5541} & \textbf{1.045} \\
\bottomrule
\end{tabular}
}
\caption{Comparison of different methods with IE-QA.}
\label{tab:tie_bench}
\end{table*}

\begin{figure}[hbp]
\centering
\includegraphics[width=1.0\columnwidth]{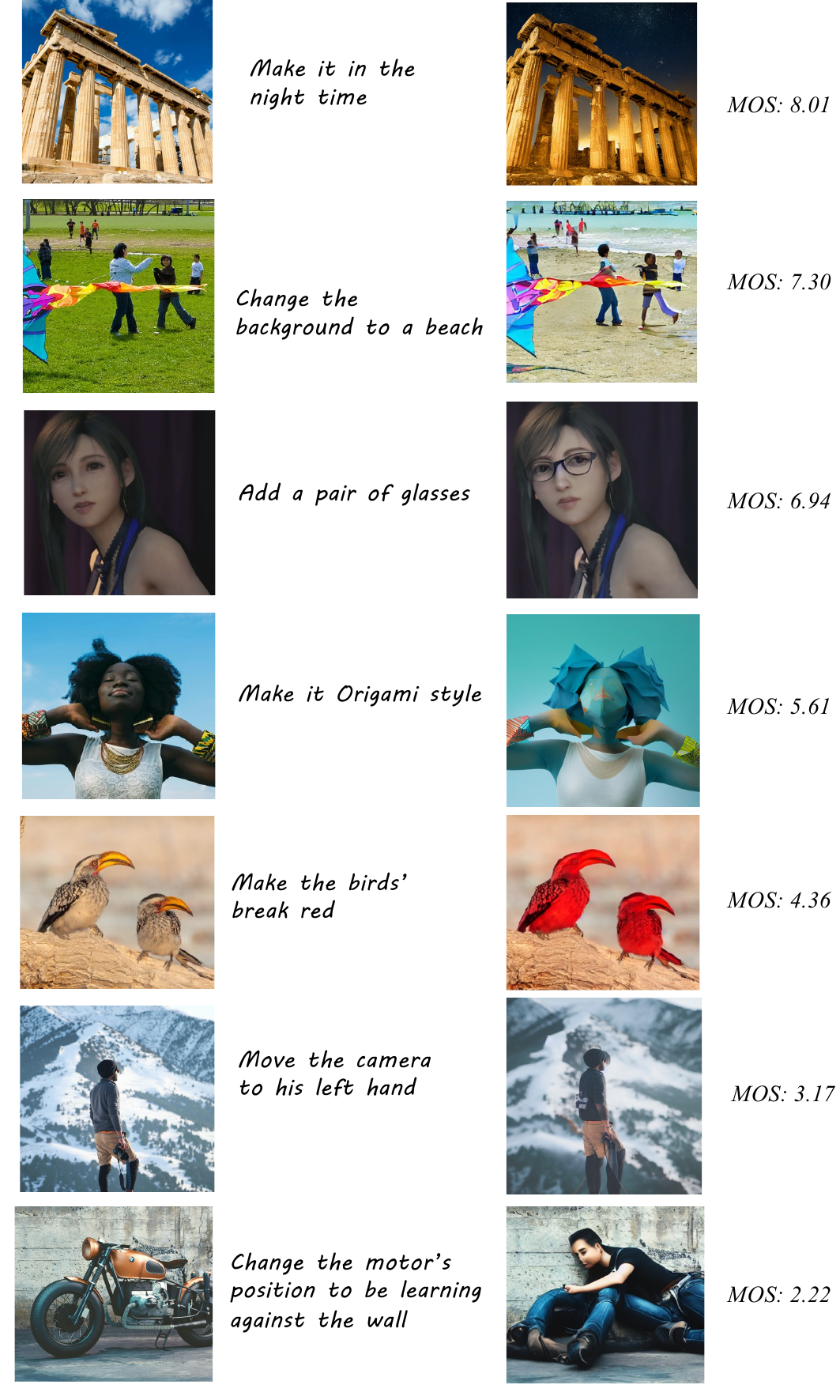} 
\caption{Demo of scores in IE-QA.}
\label{fig:demo}
\end{figure}


\begin{table*}[htb]
\centering
\scalebox{1.0}{
\begin{tabular}{llcccc}
\toprule
Experiment    & Method   & SROCC $\uparrow$ & PLCC $\uparrow$ & KRCC $\uparrow$        \\ \midrule
Baseline & IP-IQA~\cite{IP-IQA} & 0.6254 & 0.6149 & 0.5053 \\ \midrule
\multirow{2}{*}{Text-Image Alignment} 
& w/o & 0.6949 & 0.6635 & 0.5243 \\
& \underline{w} & \textbf{0.7520} & \textbf{0.7498} & \textbf{0.5541} \\ \midrule
\multirow{2}{*}{Source Connection} 
& w/o & 0.7103 & 0.7021 & 0.5195 \\
& \underline{w} & \textbf{0.7520} & \textbf{0.7498} & \textbf{0.5541} \\ \midrule
\multirow{3}{*}{Source-Target Fusion}
& Identity & 0.7103 & 0.7021 & 0.5195 \\
& Attention & 0.7422 & 0.7303 & 0.5483  \\
& \underline{Concatenation} & \textbf{0.7520} & \textbf{0.7498} & \textbf{0.5541} \\ \midrule
\multirow{2}{*}{Additional Parameters}
& w/o & 0.7354 & 0.7271 & 0.5365 \\
& w & \textbf{0.7520} & \textbf{0.7498} & \textbf{0.5541} \\ \bottomrule
\end{tabular}
}
\caption{Ablation study of the proposed IE-QA.}
\label{tab:ablation}
\end{table*}

\subsection{Qualitative Results}
We further conducted a qualitative comparison for different score levels in IE, as illustrated in Figure~\ref{fig:demo}, where we present several image examples in IE DB with varying predicted sores.

\subsection{Ablation Study}
To further validate the results of each module in IE QA, we performed detailed ablation experiments on each module, as shown in Table~\ref{tab:ablation}. All results were obtained through 10-fold validation training on the IE DB with the same experimental hyper-parameter design. The settings we adopted in our final model are underlined.
We first explore different ways of image-text alignment. Here, we experimented with CLIP and fine-tuned the regression head composed of Feed-Forward Networks, which learn the alignment from the cosine similarity of its visual and text Backbone outputs. 
Experiments demonstrate that, the text-visual branch is of importance to enhance network performance. Furthermore, we explored how to effectively model the relevance between the source image and the edited image. We first validate the effectiveness of applying source-target relationship modeling. Additionally, we explored effective ways to fuse features from the source image and the edited image, as presented in Table~\ref{tab:ablation}. ``Attention" denotes the mutli-head cross-attention. We found that concatenation along the dimension is a simple and effective design for the assessment. We further ablate the effect of additional parameter, which demonstrates the improvements are not from more parameters. In specific, the introduction of the new source branch brought additional parameters. We need to demonstrate that the model's improvement is due to the correct design (i.e., establishing a connection between the destination and source images) rather than merely from the accumulation of these additional parameters. Therefore, we conducted an experiment where we removed the source branch and instead widened the structure of the destination branch to match the parameters added by including the source branch. This adjustment helps eliminate the influence of additional parameters introduced by the source branch.
During these experiments, we could learn that the design of image-text similarity and the focus on the source-target image relationship modeling is of importance to the overall performance.


%
\section{Conclusion}
In this research, we present IE-Bench, a comprehensive framework for the evaluation of text-driven image editing techniques. This framework encompasses two key components: IE-DB and IE-QA. IE-DB is a carefully curated, subjectively aligned dataset specifically designed to address the unique challenges in assessing text-driven image editing. It includes an extensive array of video content, meticulously categorized editing prompts, and a diverse collection of edited images generated by various state-of-the-art image editing models. Notably, IE-DB stands out as the first image quality assessment (IQA) dataset that is purpose-built for evaluating the outcomes of text-driven image editing, to the best of our knowledge. Extensive experiments demonstrate the effectiveness of IE-QA. Unlike conventional metrics, IE-QA is designed to better reflect human judgment and perception, offering superior alignment with subjective evaluations of image quality and editing efficacy.

{
    \small
    \bibliographystyle{ieeenat_fullname}
    \bibliography{main}
}


\end{document}